\documentclass{article}

\usepackage{amsmath,amsfonts,bm}

\def\eqref#1{equation~\ref{#1}}

\def\ceil#1{\lceil #1 \rceil}

\def\1{\bm{1}}

\DeclareMathAlphabet{\mathsfit}{\encodingdefault}{\sfdefault}{m}{sl}
\SetMathAlphabet{\mathsfit}{bold}{\encodingdefault}{\sfdefault}{bx}{n}

\usepackage{arxiv}

\usepackage[utf8]{inputenc} 
\usepackage[T1]{fontenc}    
\usepackage{hyperref}       
\usepackage{url}            
\usepackage{booktabs}       
\usepackage{amsfonts}       
\usepackage{nicefrac}       
\usepackage{microtype}      
\usepackage{lipsum}		
\usepackage{graphicx}
\usepackage{natbib}
\usepackage{doi}
\usepackage{amsmath}
\usepackage{mathtools}

\title{Multiple Instance Learning for Brain Tumor Detection from Magnetic Resonance Spectroscopy Data}

\author{Diyuan~Lu\thanks{D. Lu and J. Triesch are with Frankfurt Institute for Advanced Studies, 60438 Frankfurt am Main, Germany. e-mail: elu, triesch@fias.uni-frankfurt.de.}%
    \And
    Gerhard~Kurz\thanks{G. Kurz is an independent researcher. e-mail: kurz.gerhard@gmail.com}%
    \And
	Nenad~Polomac\thanks{N. Polomac, I. Gacheva and E. Hattingen are with the Institute for Neuroradiology at Frankfurt university hospital, 60528 Frankfurt am Main, Germany. e-mail:Nenad.Polomac@kgu.de, elke.hattingen@kgu.de}%
	\And
	Iskra~Gacheva$^\ddagger$ \And
	Elke~Hattingen$^\ddagger$ \And
	Jochen~Triesch$^*$
	}

\date{}

\renewcommand{\shorttitle}{Multiple Instance Learning for Brain Tumor Detection from MRS Data}
\newcommand{\numspectra}{M}

\begin{document}
\maketitle

\begin{abstract}
    We apply deep learning (DL) on Magnetic resonance spectroscopy (MRS) data for the task of brain tumor detection.
    Medical applications often suffer from data scarcity and corruption by noise. Both of these problems are prominent in our data set. 
    Furthermore, a varying number of spectra are available for the different patients. We address these issues by considering the task as a multiple instance learning (MIL) problem. Specifically, we aggregate multiple spectra from the same patient into a ``bag'' for classification and apply data augmentation techniques.
    To achieve the permutation invariance during the process of bagging, we proposed two approaches: (1) to apply min-, max-, and average-pooling on the features of all samples in one bag and (2) to apply an attention mechanism.
    We tested these two approaches on multiple neural network architectures. We demonstrate that classification performance is significantly improved when training on multiple instances rather than single spectra. We propose a simple oversampling data augmentation method and show that it could further improve the performance. Finally, we demonstrate that our proposed model outperforms manual classification by neuroradiologists according to most performance metrics.
\end{abstract}

\keywords{Tumor detection \and Multiple instance learning \and Machine learning \and Magnetic resonance spectroscopy (MRS)}

\section{Introduction}
    \begin{figure}
        \centering
        \includegraphics[width=0.995\textwidth]{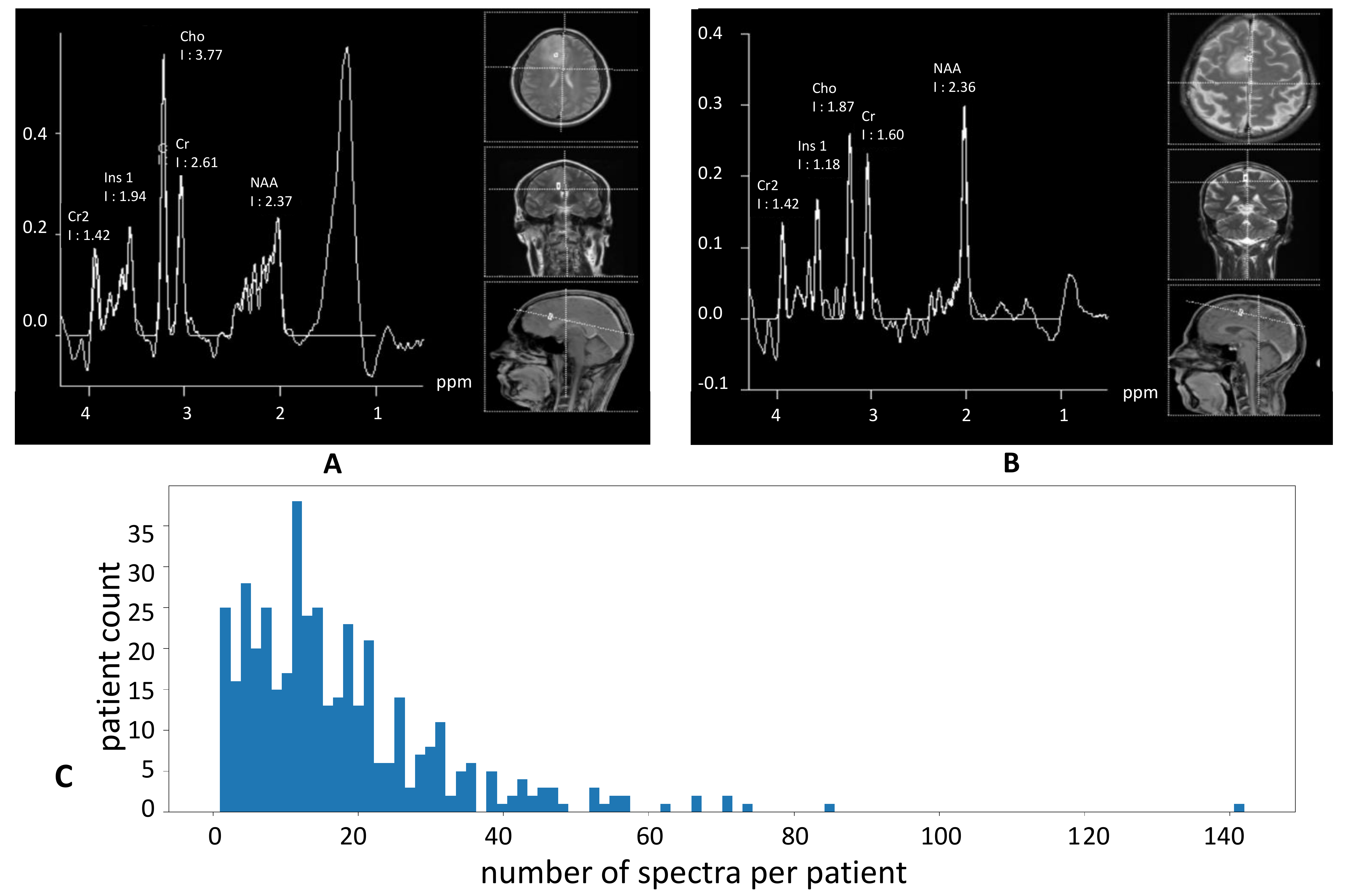} 
        \caption{Overview of the MRS data used in this paper. Each spectrum is a data array with 288 data points with the \textit{x}-axis indicating the position of different metabolites and the \textit{y}-axis indicating the intensity of the corresponding metabolites. Several spectra may stem from the same patient. \label{fig:example-hist} \textbf{A.} An example of a \textbf{tumor} MR spectrum. \textbf{B.} An example of a \textbf{non-tumor} MR spectrum. \textbf{C.} Histogram of the number of spectra per patient (17 $\pm$ 15, mean $\pm$ standard deviation). }
    \end{figure}
    
    We study the problem of brain tumor detection from MRS data. A brain tumor is the abnormal growth of the brain tissue, which can be benign or cancerous. In clinical practice, MRS is a common non-invasive tool used to identify a brain tumor, because it can be easily acquired alongside commonplace MR imaging procedures and it uniquely reflects the biochemical composition of the brain tissue {\em in situ}. MRS measures the resonant frequency shift of a chemically bound hydrogen atom (i.e., a proton), which characterizes different physiological or pathological brain metabolites. There has been increasing interest in MRS for clinical use because of the semiautomatic data acquisition, processing, and quantification~\citep{ ranjith2015machine, hatami2018magnetic, gonzalez2009using, olliverre2018generating, cruz2011semi}. However, the interpretation of MRS spectra is traditionally performed by human radiologists based on the concentration ratios of certain metabolites. In contrast, we train a model to learn informative features from the spectra as a whole.

    A common problem with MRS data is that they are often corrupted by noise from head movements during the procedure or baseline distortions of the spectrum. This poses difficulties in the MRS data interpretation. Additionally, labels are only provided per patient and not per voxel, which could introduce labeling noise as spectra from the tumor-affected hemisphere can be falsely labeled as ``tumor'' even though they contain healthy brain tissue.

    Our contributions are summarized as follows.
    \begin{itemize}
        \item We present a multiple-instance-learning (MIL)-based framework for MRS-based tumor detection that performs patient-wise classification.
        \item We propose two modules to achieve permutation invariance when processing bags of instances simultaneously, i.e., an attention module and the concatenation of max-, min-, and average-pooling, which we refer to as the ``3Pool'' module.
        \item We demonstrate that our proposed modules can be easily plugged in any given DNN-based model and improve the classification performance.
        \item We evaluate the proposed method with a leave-patient-out cross validation scheme, which carefully tests the trained model on data from unseen patients. We also show that our method is even able to outperform human neuroradiologists.
    \end{itemize}

\section{Related Work}
\label{related}
    Modern machine learning approaches based on deep neural networks (DNNs) have recently obtained impressive results in a range of classification tasks, sometimes even outperforming human experts. These successes are based on, amongst others, (1) better learning algorithms, 2) fast computing hardware, and 3) large, carefully annotated data sets. This has motivated a range of applications in oncology such as  tumor detection, tumor segmentation, tumor progression estimation\cite{lin2019super, capper2018dna, park2019machine, pereira2016brain, ranjith2015machine}, tumor grade classification~\citep{ranjith2015machine}, etc. However, acquiring the required labeled data is often hard to achieve or expensive in certain medical applications where the numbers of patients may be quite small. 
    Multiple instance learning (MIL) is a framework to handle scenarios where detailed annotations for each individual instance is noisy, laborious to obtain, or simply not available. It tries to make a decision based on a set of single instances instead of a decision for each single instance. MIL has been widely used in medical applications such as breast cancer detection \citep{sudharshan2019multiple, conjeti2017deep, sadafi2020attention} and other forms of computer assisted diagnosis \citep{fung2007multiple,liu2018landmark}. 

    Applying machine learning methods to medical applications with MRS data is gaining more and more momentum, for example in brain tumor detection \citep{gonzalez2009using, cruz2011semi, rao2015brain}, brain tumor segmentation \citep{dvovrak2015local, pereira2016brain}, breast tumor detection \citep{tavolara2019modular, ren2015faster}, and tumor motion prediction \citep{lin2019super}. There is also work to investigate the effect of the length of the echo time used to perform MR spectroscopy for the tumor detection \citep{gonzalez2009using}. \citet{olliverre2018generating} proposed to use generative-adversarial-network-based model to synthesize MRS data with real-world appearance and features for deep model training. \citet{cruz2011semi} proposed a variant of generative topographic mapping method for diagnostic discrimination between different brain tumor pathologies and the outcome prediction. 

    Noisy labels are ubiquitous in the real world. We use the term \emph{noisy labeling} to refer to annotations that are incorrect, i.e., due to the labeling procedure, the label assigned patient-wise, so they reflect the overall diagnosis rather than properties of a specific spectrum. Noisy labels are posing a non-trivial problem in deep model learning when an increasing ability to fit noise is accompanied with deeper layers. Given the ubiquity and importance of coping with noisy labeling, many works have been devoted to combating this problem. Some of them start with a small set of clean expert-labeled data~\citep{han2018co, Li2017a, VeitLearning, albarqouni2016aggnet}, but this may not be trivial to obtain. Consequently, models that can learn directly with noisy labels~\citep{han2018co, smyth2019training, RolnickDeep} are highly desirable.
    
    Multiple instance learning (MIL) is a framework to combat the problems, where detailed annotation for each single instance is noisy, or is laborious to obtain, or simply not available. Single-Instance Learning is a ``naive'' approach that assigns all instances in one bag the same label as its bag, which might lead to mislabeling negative instances in positive bags \citep{ray2005supervised}. \citet{andrews2002support} proposed to modify the standard SVM so that the MI assumption that at least one instance in each bag is positive is applicable. The normalized set kernel (NSK) and statistics kernel methods apply kernels to map the whole bags of instances into features, then use the standard SVM to make the classification on the bag level \cite{gartner2002multi}. MIL has also been widely used in medical applications such as breast cancer detection \citep{sudharshan2019multiple, conjeti2017deep}, computer assist diagnosis \citep{fung2007multiple}, brain disease diagnosis \citep{liu2018landmark}, lung cancer diagnosis \citep{ozdemir20193d}, blood cell disorder analysis \citep{sadafi2020attention}, etc. 
\section{Methods}

\subsection{Data}\label{sec:data}
    In this study, We use 1H-MR-spectroscopy data collected from 435 patients recorded in the Institute for Neuroradiology of the University Hospital in Frankfurt between 01/2009 to 3/2019. They were reviewed retrospectively and have been completely anonymized for this study. The patients were suffering from either glial or glioneuronal first diagnosed tumors (the \textit{tumor} group, 266 patients) or other non-neoplastic
    lesions, e.g., demyelination, gliosis, focal cortical dysplasia, enlarged Virchow-Robin spaces or similar (the \textit{non-tumor} group, 156 patients). The tumor group included all spectra from the tumor-affected hemisphere. The non-tumor group consisted of spectra from both hemispheres of the patients. 
    
    As a result, 7442 spectra (3388 non-tumor and 4054 tumor) were selected for further analysis. The obtained MRS examples are saved as 1-\textit{d} arrays with 288 data points, i.e., in shape ($288\times1$), shown in Fig.~\ref{fig:example-hist}A, B, 
    where the $y$-axis shows signal intensities of different metabolites, and the $x$-axis represents the chemical shift positions in ppm indicating various metabolites. The indices correspond to the position of metabolites and the values indicate signal intensities of corresponding metabolites.
    We normalize each spectrum to zero mean and unit variance. All spectra from the same patient are labeled with the patient's diagnosis, i.e., all spectra from one tumor patient will be labeled as \textit{tumor}, and all spectra from one non-tumor patient would be labeled as \textit{non-tumor}.
    
    
    %
    There is a huge variance in the number of spectra per patent in our data set - some patients have dozens of spectra and some have just a few spectra or even just a single one. A histogram of the number of spectra per patient is shown in Fig.~\ref{fig:example-hist}C. 
    Due to the fairly limited number of patients, machine learning methods trained on this data set are prone to overfitting, therefore applying out-of-the-box methods would not yield satisfactory results.
    Each spectrum describes the biochemical composition of one voxel of brain tissue. We propose to perform classification not on a single spectrum, but on a bag of spectra from this patient. Specifically, we create bags of spectra from each patient for training and validation.

    \begin{figure}
        \centering
        \includegraphics[width=0.8\textwidth]{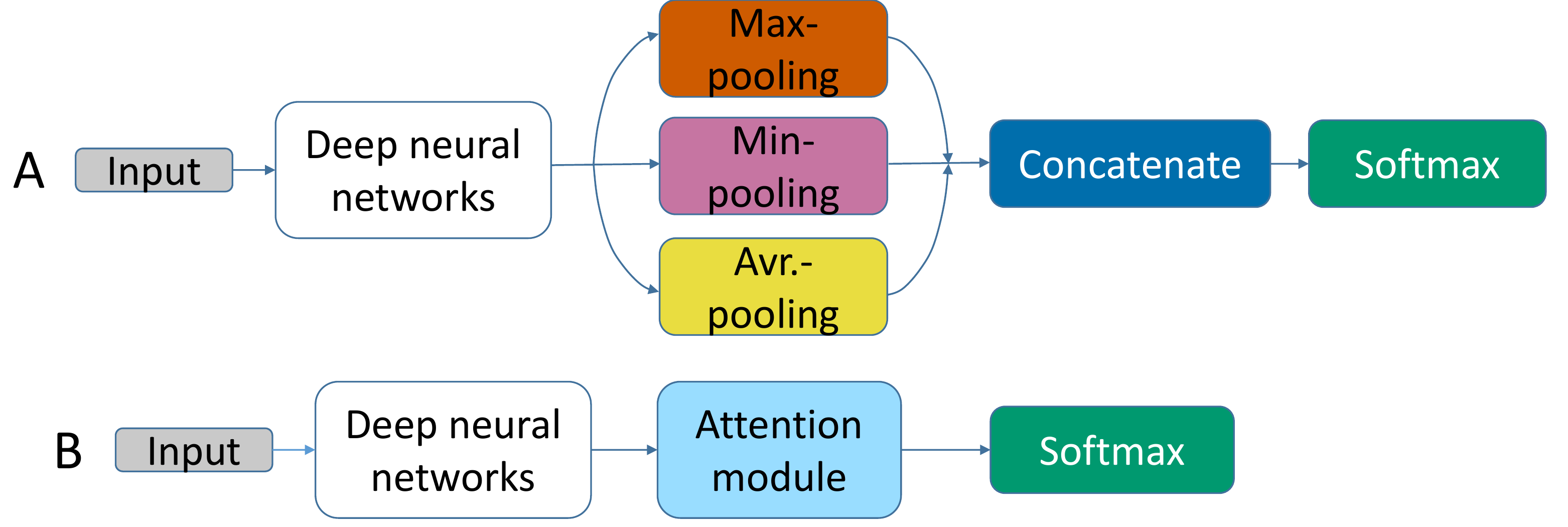}
        \caption{\label{fig:structure} Overview of the proposed framework with two proposed permutation invariant modules, which can be plugged in any DNN-based models. \textbf{A.} The proposed ``3Pool'' module. \textbf{B.} The proposed attention module.}
    \end{figure}

\subsection{Patient-wise Data Preparation}
\label{bags-generation}
    We have MRS spectra from a total number of $P$ patients, the total number of spectra for patient $p$ is $N_p$. We generate bags of spectra consisting of a fixed number $\numspectra \in \mathbb{N}$ of spectra from each patient by sampling from all spectra of the patient with replacement during training. Each bag is in shape $\numspectra \times 288$. The bags from patient $p$ are denoted as $\textbf{X}^p = \{ \textbf{x}_1^p, \dots, \textbf{x}_{p_b}^p\}$, where $p_b$ is the total number of bags generated for patient $p$. Since, this is a combinatorial problem, we could potentially generate millions of samples. This could be viewed as a data augmentation (DA) process. However, the more bags we generate from one patient, the less diversity we introduce through the DA and the worse the network is at generalization. Empirically, we set the number of generated bags of one patient to three times their single spectra count. Of course, further exploration of the optimum number of spectra to use might be beneficial in the future. Each training bag is provided a class label $y^p \in \{\textit{tumor}, \textit{non-tumor} \}$ based on the diagnosis of the patient. 
    More formally, our goal is to learn a function \textit{f}, which takes a set of spectra $\textbf{x}^p = \{x^p_i, \dots, x^p_N\}$ from patient $p$, and output the classification decision $\hat{y}^p$.
    The function \textit{f} processes all spectra at the same time and generates a final predicted label $\hat{y}$. The training objective is the classic cross entropy loss: 
    \begin{equation}
        \underset{\theta}{\mathrm{min}}~\mathbb{E}_{P(\textbf{x}, \hat{y})} [-\log P_{\theta} (y = \hat{y}|\textbf{x})],
    \end{equation}
    where $\theta$ refers to the parameters of the function \textit{f}.
    
    The ability of the classifier to generalize to new previously unseen patients is of great clinical importance. Therefore, we apply a 5-fold leave-subjects-out cross validation scheme. To be specific, we divide the patient list into 5 sub-lists, each with around 80 patients. In each cross validation set, we withhold the data from the patients of one sub-list, while we train and validate on the data from the other sub-lists. During training and validation, we adopt a 4:1 split ratio of all generated bags. During testing, we switch off the data augmentation strategy and only allow the minimal repetition of the spectra to fill up the last bag, which may be only partially filled otherwise. This makes sure that the number of bags to generate for patient $p$ follows 
\begin{equation}
    p_b  =
    \begin{cases}
            1, &         \text{if } N_p \leq \numspectra,\\
            \ceil{N_p/ \numspectra}, &         \text{if } N_p > \numspectra.
    \end{cases}
\end{equation}

\subsection{Network Structure}
    When working with bags of MRS spectra, we note that the order of the stacked spectra was randomly chosen and should not affect the result of the network. Being invariant to the order of the spectra can either be achieved by augmenting with shuffled data, which is an approximation, or by designing the network architecture in such a way that the output of the network is independent of the order of the spectra in the input. In this work, we compared both approaches. For the former, we have described the data augmentation that we use to generate bags of training samples in section \ref{bags-generation}. For the latter, we proposed two modules that can be easily plugged in any DNN-based models: (1) to aggregate the minimum-, maximum- and mean-pooling of the feature maps which yields exact order invariance, (2) to leverage attention mechanism \citep{ilse2018, sadafi2020attention}, where different instances in the bag are assigned with different attention weights, which can be learned by the neural network. The schematic of propose method is shown in Fig.~\ref{fig:structure}. The final extracted feature is a weighted average of features from all the instances in one bag. Since the attention weights depend on the instance itself and not the order, we can also achieve exact permutation invariance.
    
    In this work, we test the two proposed modules on several network structures, i.e., a multi-layer perceptron  (MLP), an Inception-variant tailored to MRS data, and a CNN model inspired by Hatami \textit{et al.}~\cite{hatami2018magnetic}. An Inception model is a successful neural network structure proposed to scale up convolution networks in efficient ways \cite{szegedy2016rethinking}. In our implementation, we only preserve the first five inception blocks from the original InceptionV3 model \cite{szegedy2016rethinking} and reduce the number of filters in each block compared to the original configuration due to a lower complexity of our MRS data compared to the image data. In the MLP model, there are three dense layers with 128, 32, and 2 dense units, respectively, as shown in Fig.~\ref{fig:structure}B. In the model inspired from Hatami \textit{et al.}, we omit the last convolutional layer with 512 kernels and the max-pooling layer, since the length of our data is smaller than theirs. 
    Furthermore, for each model, we consider two variants, i.e., the one with concatenated max-, min-, and average-pooling, denoted ``3Pool'' and the other with an attention module, denoted ``Att''. Note that the feature extraction in each dense layer is performed on the single instance level, i.e., the convolution is only done horizontally with the kernel height as one. The feature maps are then either pooled and concatenated in a ``3Pool'' branch, or processed by the attention module.
    

\subsection{Attention Module}

    In order to weigh the different samples contained in a bag, we make use of the attention mechanism proposed by \cite{ilse2018}. The idea is to introduce a layer whose output $z$ is a weighted average
    \begin{align}
        z = \sum_{k=1}^\numspectra a_k h_k
    \end{align}
    of the inputs $h_k$ with weights
    \begin{align}
        a_k = \frac{\exp\left(w^T \tanh(V h_k^T) \right)}{
        \sum_{k=1}^\numspectra \exp\left(w^T \tanh(V h_k^T) \right)} \ ,
    \end{align}
    where $w \in \mathbb{R}^{1 \times N_{att}}$ and $V \in \mathbb{R}^{N_{att} \times L_{h_k}}$ are learned parameters of the layer. $N_{att}$ is the number of attention heads and $L_{h_k}$ is the dimension of the hidden feature $h_k$. As each $a_k$ depends on the values inside $h_k$, the weights are different in each bag and can take the concrete values inside the input bag into account. Note that the output $z$ is independent of the order of the inputs $h_k$ 
 
 \subsection{Training Procedure}\label{sec:training}
    The network is trained with randomly initialized weights using the Adam optimizer with default parameters and a mini-batch size of 32. The model is trained on a Windows machine with an Intel(R) Core i7-4770 CPU, 16 GB RAM and a GeForce GTX1060 GPU with 6GB of memory. The training and takes less than 3 minutes for 30 training epochs.

\section{Results}
\label{results}
	
\begin{table}[t]
    \caption{Performance matrices averaged across \textbf{five-fold cross validation data sets }of proposed method compared to other baseline methods. The results are shown in $\textbf{mean} \pm \textbf{standard deviation}$. MCC: Matthews correlation coefficient, AUC: area under ROC curve. SI: single instance. Baseline MIL models: MI-SVM \citep{andrews2002support},  Ray-MISVM \citep{ray2005supervised}, and NSK \citep{gartner2002multi}. MLP: multi-layer perceptron.}
    \label{tab:performance}
    \begin{center}
    \begin{tabular}{lccccc}
    \toprule
    &  \textbf{Bag} &  \textbf{Patient}&  &   \\
   &\textbf{AUC} &  \textbf{AUC}  & \textbf{F1-score}& \textbf{MCC} & \textbf{\# Trainables} \\
   \midrule
    Ray-MISVM \cite{ray2005supervised} (SI) &$0.73 \pm 0.07$ & $0.74 \pm 0.06$  & $0.69 \pm 0.06 $ & $0.31 \pm 0.12$ &$\sim$600\\
    Ray-MISVM \cite{ray2005supervised} &$0.63 \pm 0.02$ & $0.59 \pm 0.02$  & $0.54 \pm 0.03 $ & $0.19 \pm 0.05$ &$\sim$600\\
    Ray-MISVM \cite{ray2005supervised} + DA &$0.73 \pm 0.09$ & $0.73 \pm 0.08$  & $0.71 \pm 0.10 $ & $0.35 \pm 0.18$ &$\sim$600\\
    \midrule
	MI-SVM \cite{andrews2002support} (SI)   & $0.71 \pm 0.04$ & $0.74 \pm 0.06$  & $0.68 \pm 0.04 $ & $0.30 \pm 0.09$ &$\sim$600\\
	MI-SVM\cite{andrews2002support}  & $0.69 \pm 0.07$ & $0.69 \pm 0.07$  & $0.69 \pm 0.07 $ & $0.30 \pm 0.12$ &$\sim$600\\
    MI-SVM\cite{andrews2002support} + DA  & $0.69 \pm 0.08$ & $0.69 \pm 0.07$  & $0.70 \pm 0.07 $ & $0.30 \pm 0.12$ &$\sim$600\\
    \midrule
    NSK \cite{gartner2002multi} (SI) & $0.72 \pm 0.05$ & $0.72 \pm 0.05$  & $0.71 \pm 0.05 $ & $0.34 \pm 0.09$  &$\sim$600\\
    NSK \cite{gartner2002multi}  &$0.70 \pm 0.06$ & $0.69 \pm 0.06$  & $0.69 \pm 0.04 $ & $0.30 \pm 0.11$ &$\sim$600\\
    NSK \cite{gartner2002multi} + DA & $0.74 \pm 0.04$ & $0.74 \pm 0.04$  & $0.72 \pm 0.03 $ & $0.35 \pm 0.06$ &$\sim$600\\
	\midrule		
	MLP (SI) & $0.73 \pm 0.04$ & $0.77 \pm 0.05$  & $0.69 \pm 0.05 $ & $0.30 \pm 0.08$&41,314\\
	MLP-3Pool  & $0.68 \pm 0.07$ & $0.68 \pm 0.08$  & $0.69 \pm 0.04 $ & $0.30 \pm 0.09$  &41,314\\
	MLP-3Pool + DA  & $0.72 \pm 0.11$ & $0.72 \pm 0.10$  & $0.70 \pm 0.07 $ & $0.33 \pm 0.15$ &41,314\\
	MLP-Att & $0.78 \pm 0.08$ & $0.78 \pm 0.08$  & $0.73 \pm 0.08 $ & $0.37 \pm 0.17$ & 41,220\\
	MLP-Att + DA& $ 0.79 \pm  0.06 $ & $ 0.79  \pm  0.05 $  & $ 0.74  \pm  0.05  $ & $ 0.42  \pm  0.11 $& 41,220\\
    \midrule
	Hatami (SI)  & $0.67 \pm 0.03$ & $0.72 \pm 0.04$  & $0.65 \pm 0.03 $ & $0.23 \pm 0.06$ &488,514\\
	Hatami-3Pool  & $0.77 \pm 0.07$ & $0.76 \pm 0.06$  & $0.72 \pm 0.05 $ & $0.36 \pm 0.11$ &488,514\\
    Hatami-3Pool + DA   & $\textbf{0.82} \pm \textbf{0.07}$ & $\textbf{0.82} \pm \textbf{0.06}$  & $\textbf{0.78} \pm \textbf{0.08} $ & $\textbf{0.46} \pm \textbf{0.19}$ &488,514\\
	Hatami-Att   & $0.80 \pm 0.05$ & $0.80 \pm 0.04$  & $0.73 \pm 0.05 $ & $0.37 \pm 0.14$ &507,012\\
    Hatami-Att + DA & $0.81 \pm 0.06$ & $0.81 \pm 0.05$  & $0.75 \pm 0.08 $ & $0.43 \pm 0.16$ &507,012\\
    \midrule
	Inception-3Pool (SI)  & $0.73 \pm 0.07$ & $0.76 \pm 0.07$  & $0.69 \pm 0.07 $ & $0.32 \pm 0.14$ &345,098\\
	Inception-3Pool  & $0.77 \pm 0.07$ & $0.76 \pm 0.06$  & $0.72 \pm 0.05 $ & $0.36 \pm 0.11$ &345,098\\
    Inception-3Pool + DA & $ 0.79  \pm  0.05 $ & $ 0.79  \pm  0.05 $  & $ 0.74  \pm 0.05  $ & $0.39 \pm 0.09$&345,098\\
	Inception-Att  & $0.75 \pm 0.06$ & $0.75 \pm 0.05$  & $0.72 \pm 0.05 $ & $0.36 \pm 0.10$ &345,116\\
    Inception-Att + DA & $0.76 \pm 0.07$ & $0.75 \pm 0.07$  & $0.73 \pm 0.05 $ & $0.38 \pm 0.11$ &345,116\\
    \bottomrule
    \end{tabular}
    \end{center}
    \end{table}

\subsection{Overall Performance with Ablation}
    To evaluate performance, we use the area under the receiver operating characteristic (ROC) curve, the F1-Score and the Matthews correlation coefficient (MCC). The ROC curve is constructed by varying the classification threshold and calculating the true positive (TP), false positive (FP), true negative (TN), and false negative (FN) rates. We report classification accuracy, area under the ROC curve (AUC), F1-score = $ \frac{\text{2TP}}{\text{2TP} + \text{FP}+ \text{FN}}$, and MCC = $ \frac{\text{TP}\times\text{TN} - \text{FP}\times\text{FN}}{\sqrt{(\text{TP} + \text{FP})(\text{TP} + \text{FN})(\text{TN} + \text{FP})(\text{TN} + \text{FN})}}$. The MCC is generally considered as a balanced measure which takes into account \text{TP}, \text{TN}, \text{FP}, and \text{FN}, and it can be used even if the classes are not balanced.  We also conducted ablation studies on the effectiveness of data augmentation on different network structures. Moreover, we  compared our method to three other baseline methods, i.e., the support vector machine approaches by Ray-MISVM \cite{ray2005supervised}, MI-SVM \cite{andrews2002support}, and  NSK \cite{gartner2002multi}. For this purpose, we used the implementation from \citet{MISVM}.

    Empirically, we found that using 31 spectra per bag yields relatively good results. Therefore, we report the averaged performance metrics with the default $\numspectra = 31$ across all cross validation sets. The results averaged across all leave-subjects-out cross validation sets are shown in Table~\ref{tab:performance}. In addition to the comparison on multiple instances learning, we also ran all the models (1) with single instances, denoted with ``(SI)'', (2) with the oversampling data augmentation, denoted with `` + DA''. From Table~\ref{tab:performance}, we made the following observations and possible explanations.  Firstly, the CNN network inspired from Hatami \textit{et al.} with the proposed ``3Pool'' module achieved the best results: a bag AUC of 0.82, a patient-wise AUC of 0.82, an F1-score of 0.78 and an MCC of 0.46. Secondly, when grouping multiple instances into bags for training without any data augmentation, models with low complexity, indicated by the number of trainable parameters, show a performance deterioration and models with a large number of trainable parameters still show an improvement in the performance. One contributing factor might be that the number of total training samples are significantly reduced when changing from the SI learning case to MI learning, thus the generalization ability is not fully explored. Thirdly, the ``3Pool'' module works the best with high complexity models such as Hatami-model and Inception. Thirdly, data augmentation (`` + DA'') almost always helps improve the performance, except in the case of MI-SVM.  
    Thirdly, of the two proposed approaches to achieve permutation invariance, i.e., (1) using max-, min-, and average-pooling of feature maps before the softmax activation, and (2) the attention-weighted average of feature maps before the softmax activation, we found that the first approach works better when combined with the Inception network, but the second approach is superior when using the MLP. Thus, neither approach is clearly superior and the choice of method needs to be made depending on the particular structure of the underlying neural network.

\subsection{Human vs. Machine}
	We compared the performance of implemented DNN models to that of human neuroradiologists on one randomly selected test set, which has 844 spectra from around 42 patients. The result is shown in Tab.~\ref{tab:comparison}. For the collection of the classification results of neuroradiologists, we divided the test set into eight subsets and each subset was assigned to one of eight neuroradiologists. The neuroradiologists' performance therefore represents the collective effort of eight individuals, which is faithfully reflect the clinical practice. The data shows that the performance of our proposed method is better on almost all performance metrics except the MCC. The reason is that the neuroradiologists achieved a specificity of 0.88 but at a cost of a low sensitivity of 0.54. This may reflect that neuroradiologists assign different ``costs'' to false positive vs.\ false negative classifications.
\begin{table}[t]
    \caption{Performance on \textbf{withheld neuroradiologist-labeled data set} of all models. SIC: single-instance classification. MCC: Matthews correlation coefficient, AUC: area under ROC curve. SI: single instance.  Baseline MIL models: MI-SVM \citep{andrews2002support},  Ray-MISVM \citep{ray2005supervised}, and NSK \citep{gartner2002multi}}
    \label{tab:comparison}
    \begin{center}
    \begin{tabular}{lcccc}
    \toprule
    &  \textbf{Bag} &  \textbf{Patient}&  &   \\
   &\textbf{AUC} &  \textbf{AUC}  & \textbf{F1-score}& \textbf{MCC} \\
	\midrule		
    Neuroradiologists & -- & --  & 0.56  & \textbf{0.58} \\
    \midrule
    Ray-MISVM \cite{ray2005supervised} (SI) &$0.64 \pm 0.04$ & $0.60 \pm 0.03$  & $0.52 \pm 0.03 $ & $0.14 \pm 0.06$\\
    Ray-MISVM \cite{ray2005supervised} &$0.62 \pm 0.02$ & $0.59 \pm 0.02$  & $0.55 \pm 0.04 $ & $0.12 \pm 0.08$\\
    Ray-MISVM \cite{ray2005supervised} + DA &$0.63 \pm 0.02$ & $0.59 \pm 0.02$  & $0.55 \pm 0.05 $ & $0.16 \pm 0.10$\\
    \midrule
	 MI-SVM \cite{andrews2002support} (SI) & $0.67 \pm 0.02$ & $0.65 \pm 0.04$  & $0.58 \pm 0.05 $ & $0.29 \pm 0.08$\\
	 MI-SVM \cite{andrews2002support} & $0.63 \pm 0.03$ & $0.59 \pm 0.03$  & $0.60 \pm 0.04 $ & $0.26 \pm 0.09$\\
     MI-SVM \cite{andrews2002support} + DA & $0.65 \pm 0.02$ & $0.60 \pm 0.03$  & $0.62 \pm 0.03 $ & $0.26 \pm 0.05$\\
    \midrule
    NSK \cite{gartner2002multi} (SI) & $0.70 \pm 0.02$ & $0.68 \pm 0.03$  & $0.58 \pm 0.02 $ & $0.27 \pm 0.03$ \\
    NSK \cite{gartner2002multi} &$0.69 \pm 0.05$ & $0.65 \pm 0.06$  & $0.60 \pm 0.05 $ & $0.23 \pm 0.09$\\
    NSK\cite{gartner2002multi} + DA & $0.73 \pm 0.05$ & $0.69 \pm 0.05$  & $0.66 \pm 0.06 $ & $0.35 \pm 0.10$\\
    \midrule
	MLP-3Pool (SI)  & $0.74 \pm 0.06$ & $0.74 \pm 0.06$  & $0.61 \pm 0.07 $ & $0.32 \pm 0.12$ \\
	MLP-3Pool & $0.77 \pm 0.04$ & $0.70 \pm 0.05$  & $0.68 \pm 0.05 $ & $0.38 \pm 0.11$ \\
    MLP-3Pool + DA & $0.75 \pm 0.05$ & $0.69 \pm 0.06$  & $0.67 \pm 0.03 $ & $0.35 \pm 0.08$\\
	MLP-Att & $0.76 \pm 0.04$ & $0.70 \pm 0.04$  & $0.65 \pm 0.02 $ & $0.33 \pm 0.04$\\
    MLP-Att + DA & $0.78 \pm 0.03$ & $0.72 \pm 0.03$  & $0.65 \pm 0.02 $ & $0.33 \pm 0.06$\\
    \midrule
	Hatami (SI)   & $0.66 \pm 0.02$ & $0.69 \pm 0.04$  & $0.53 \pm 0.02 $ & $0.22 \pm 0.03$ \\
	Hatami-3Pool   & $\textbf{0.86} \pm \textbf{0.02}$ & $\textbf{0.80} \pm \textbf{0.03}$  & $0.74 \pm 0.03 $ & $0.49 \pm 0.06$ \\
    Hatami-3Pool + DA  & $0.84 \pm 0.02$ & $0.78 \pm 0.03$  & $0.70 \pm 0.01 $ & $0.43 \pm 0.03$\\
	Hatami-Att   & $0.83 \pm 0.04$ & $0.78 \pm 0.04$  & $0.71 \pm 0.03 $ & $0.45 \pm 0.05$\\
    Hatami-Att + DA   & $0.85 \pm 0.02$ & $0.80 \pm 0.03$  & $0.74 \pm 0.02 $ & $0.49 \pm 0.03$\\
    \midrule
	Inception-3Pool (SI)  & $0.73 \pm 0.06$ & $0.70 \pm 0.05$  & $0.61 \pm 0.06 $ & $0.34 \pm 0.10$ \\
	Inception-3Pool  & $ 0.83 \pm  0.04$ & $ 0.77 \pm  0.04$  & $ \textbf{0.75} \pm  \textbf{0.02} $ & $ 0.56\pm 0.05$ \\
    Inception-3Pool + DA & $0.81 \pm 0.04$ & $0.74 \pm 0.04$  & $0.70 \pm 0.05 $ & $0.43 \pm 0.10$ \\
	Inception-Att & $0.82 \pm 0.04$ & $0.76 \pm 0.04$  & $0.74 \pm 0.06 $ & $0.50 \pm 0.11$ \\
    Inception-Att + DA & $0.79 \pm 0.03$ & $0.74 \pm 0.04$  & $0.70 \pm 0.03 $ & $0.43 \pm 0.05$ \\
    \bottomrule
    \end{tabular}
    \end{center}
    \end{table}
    

\subsection{Attention Visualization}
\begin{figure}[tb]
        \centering
        \includegraphics[width=1.05\textwidth]{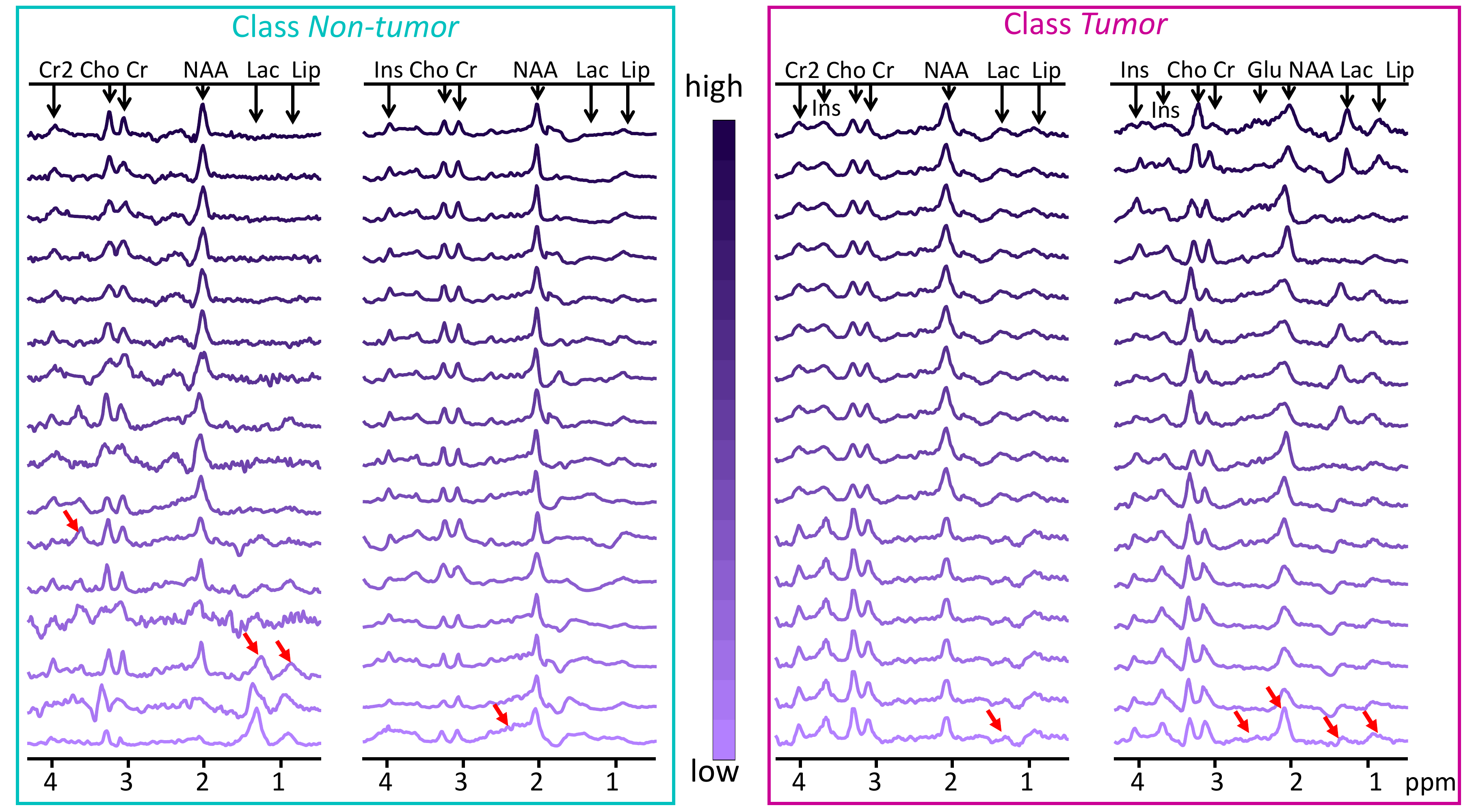}
        \caption{\label{fig:attention} Exemplar bags (column) of MRS spectra with attention, color-coded with shades in a descending order. Red arrows in instances with relatively low attention show the features from the opposite class. }
\end{figure}

Further, we show two bags of samples from each class with color-coded attention during testing, shown in Fig~\ref{fig:attention}. We can see that features of spectra in one bag are very heterogeneous exhibiting different peak ratios, peak positions, etc. Note that, samples with high attention might be stereotypical of that class or raising a red flag for that class decision. One benefit of visualizing the attention assignment is that it provides not only a final classification result but also the contextual information of the same patient's brain tissue. This could provide more information for the MRS data interpretation. The common metabolites from left to right in our data are creatine2 (Cr2, 3.9 ppm), myo-inositol and glycine (MI/Gly, $\sim$ 3.5 ppm), Myo-inositol (Ins, 3.61 ppm), choline (Cho, 3.19 ppm), creatine (Cr, 3.03 ppm), Glutamin (Glu, 2.2 -- 2.4 ppm), N-acetyl aspartate (NAA, 2.01 ppm), lactate (Lac, 1.4 ppm), and Lipids (Lip, 0.9 ppm) \cite{faghihi2017magnetic, FanIn, Fan2006Magnetic, RaeRE, hattingen20091}. There are several indicative features in MRS data that are clinically relevant. For example, in tumor spectra, there are weakened Cr and Ins \cite{faghihi2017magnetic}, reduced NAA concentration \cite{faghihi2017magnetic}, elevated Cho, Glu, Lac, Lip peaks \cite{RaeRE, Fan2006Magnetic, faghihi2017magnetic}, elevated MI/Gly \cite{hattingen20091}. 

 
In Fig.~\ref{fig:attention}, we can see that in the \textit{non-tumor} group, the high attention weights are assigned to samples with flat Lip, flat Lac \cite{RaeRE}, high and narrow NAA (low 2.0 -- 2.5 ppm), clear Cr/Cho ratio $>$ 1, etc. For the \textit{tumor} group, the high attention weights are often assigned to instances with low NAA with elevated Glu, high Lac, high Lip, clear Cr/Cho ratio $<$ 1, as shown in \cite{RaeRE, Fan2006Magnetic, hattingen20091, faghihi2017magnetic} 
    
\subsection{Varying the Bag Size}
\begin{figure}[tb]
        \centering
        \includegraphics[width=1.05\textwidth]{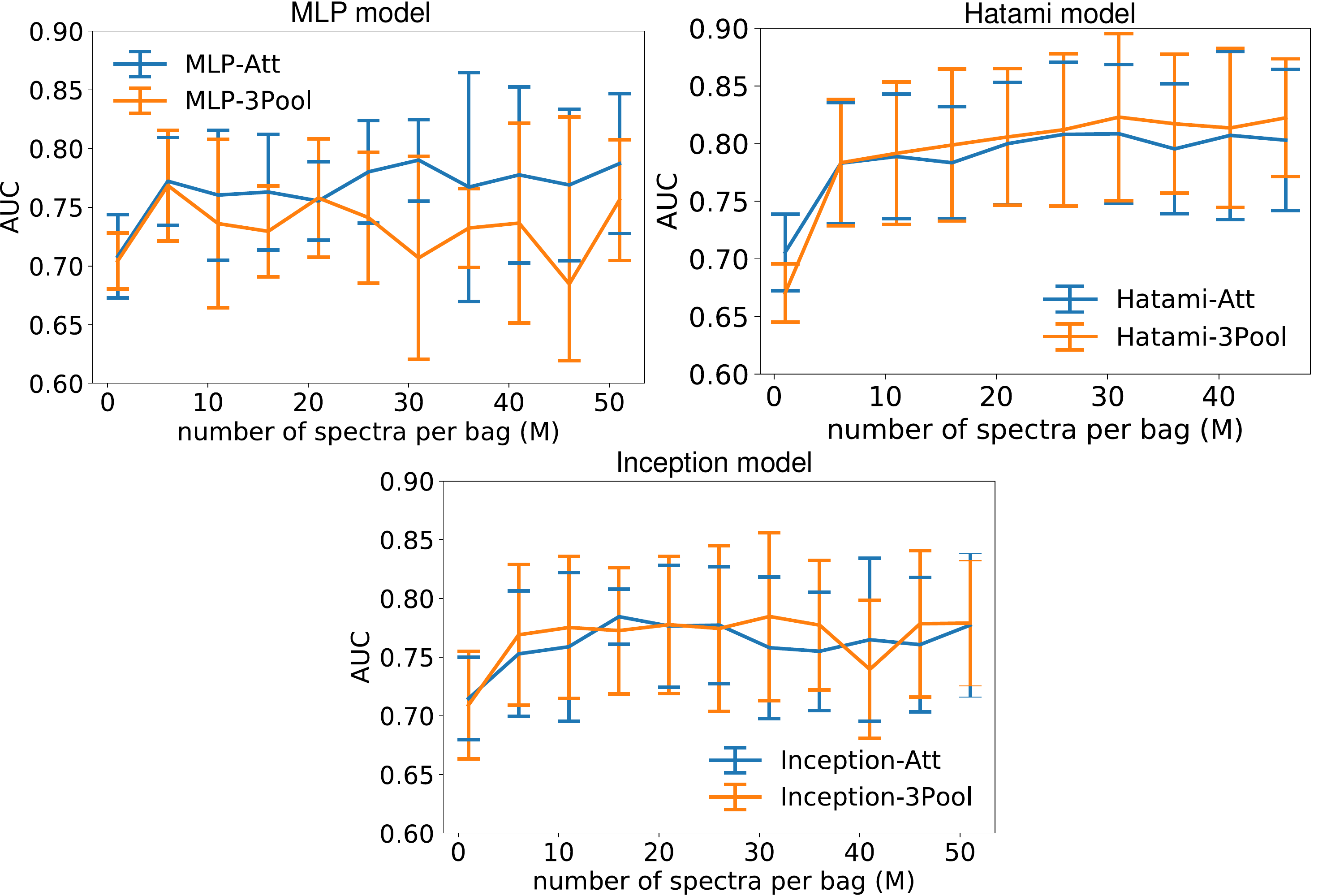}
        \caption{\label{fig:auc-all} Averaged ROC-AUC as a function of the number of instances per bag across five leave-patients-out cross validation sets for our proposed methods. The errorbars represent one standard deviation.}
\end{figure}
To investigate the effect of the number of samples per bag, we vary the value from one (corresponding to single instance classification) to 51. The AUC as a function of the number is shown in Fig~\ref{fig:auc-all}. From this experiment, we made the following observations. Firstly, for all models, learning from the bags of multiple instances is better than learning from a single instance. The performance is significantly improved when $\numspectra$ increases from one to six, and then this improvement attenuated after $\numspectra = 6$ in all models. Secondly, the performance with the attention module did not show a deterioration with an increasing $\numspectra$ in all models. However, in the MLP model, the performance degraded after $\numspectra = 6$ with the ``3Pool'' module.   


\section{Conclusion}
\label{conclusion}


This paper presents a novel framework for tumor detection based on multiple instance (MI) learning with noisily-labeled MRS data. 
We proposed two modules to achieve permutation-invariance within each bag: (1) an attention module and (2) a ``3Pool'' module with max-, min-, and average-pooling. Moreover, we applied data augmentation to generate bags of instances from each patient, which expanded the total training data size as well as increased the variance in the training data. We applied these two modules on several popular DNN models, i.e., an MLP, an Inception-variant, and a CNN-based model inspired by Hatami et al. \citet{hatami2018magnetic}. We conducted a thorough comparison between the different models as well as three conventional SVM-based MI methods. We also carried out an ablation study regarding the effect of the data augmentation for all models. We observed the MI SVMs do not perform well on our data set. 
The data augmentation almost always improved the performance compared to the counterpart without augmentation, except in the case of \citet{andrews2002support}.
In the Hatami-model and the Inception model, the proposed ``3Pool'' module achieved slightly better performance than the ``Att'' module. However, in the MLP model, the proposed ``Att'' module was superior. 
The best results of all experimented configurations were obtained by the Hatami-model with the proposed ``3Pool'' module and data augmentation: a bag AUC of 0.82, a patient-wise AUC of 0.82, an F1-score of 0.78, and an MCC of 0.46. We showed that our MI-based approach significantly improved the performance compared to single instance classification (t-test with a p-value of $\leq$ 0.004) and that applying data augmentation for generating more training data is beneficial to obtain good results, however it does not rise to the level of being statistically significant. We also demonstrate that the proposed method outperforms human radiologists in terms of F1-score while achieving a similar MCC.
The limitation of this work is that the results are obtained from a data set collected from a single site. Due to the factors such as the variability of data acquisition procedures, the diverse patient populations, the generalization ability of the proposed method to other MRS data sets is not demonstrated. Furthermore, so far we only experimented with a very simple data augmentation method. Further exploration of other data augmentation strategies such as mixup, adding noise, scaling amplitude, etc., might be interesting in the future. 
A further inspection of the different effects of ``Att'' and ``3Pool'' to the learning of different networks is also of interest. So far, we used a stratified sampling strategy, i.e., the more single spectra one patient has, the more bags we generate. This could potentially introduce bias. In the future, we could fix the number of bags to generate for all patients to eliminate the bias introduced by the current method. Furthermore, we could explore other statistics within the bag such as the median and the interquartile range. Adding explainable machine learning methods is also beneficial for promoting the approach for clinical practice. Finally, we would like to investigate the behaviour of the proposed approaches on further data sets collected at other sites.


\bibliography{references}
\bibliographystyle{abbrvnat}

\end{document}